\documentclass[11pt,a4paper]{article}
\usepackage[hyperref]{acl2020}
\usepackage{times}
\usepackage{latexsym}

\usepackage{microtype}

\usepackage{enumitem}
\usepackage{amsmath}
\usepackage{amsfonts}

\usepackage{spverbatim}
\usepackage[linesnumbered,ruled,vlined]{algorithm2e}
\usepackage{graphicx}
\usepackage{lscape}
\usepackage{longtable}

\usepackage{booktabs}
\usepackage{multirow}
\usepackage{graphicx}
\usepackage{xcolor}
\usepackage{makecell}

\DeclareMathOperator*{\argmax}{argmax}

\aclfinalcopy % Uncomment this line for the final submission
\title{How to Evaluate Word Representations of Informal Domain?}

\author{Yekun Chai \\
  University of Edinburgh \\
  \texttt{chaiyekun@gmail.com} \\\And
  Naomi Saphra \\
  University of Edinburgh \\
  \texttt{n.saphra@ed.ac.uk} \\
  \And
  Adam Lopez \\
  University of Edinburgh \\
  \texttt{alopez@inf.ed.ac.uk} \\}
\date{}

\begin{document}
\maketitle
\begin{abstract}
Diverse word representations have surged in most state-of-the-art natural language processing (NLP) applications. Nevertheless, how to efficiently evaluate such word embeddings in the informal domain such as Twitter or forums, remains an ongoing challenge due to the lack of sufficient evaluation dataset. We derived a large list of variant spelling pairs from \emph{UrbanDictionary} with the automatic approaches of weakly-supervised pattern-based bootstrapping and self-training linear chain conditional random field (CRF). With these extracted relation pairs we promote the odds of eliding the text normalization procedure of traditional NLP pipelines and directly adopting representations of non-standard words in the informal domain. Our code is available\footnote{\url{https://github.com/cyk1337/UrbanDict}}. 

\end{abstract}

\section{Introduction}
% bg
Distributional word representation is an impressive approach to denoting the natural language words with low-dimensional real-valued vectors that could implicitly signal the syntactic or semantic statistics of corresponding words, which has been extensively used in amounts of current NLP systems, \emph{e.g.} text classification~\citep{kim2014convolutional,yang2016hierarchical,joulin2016bag, xiang2019adaptive}, name entity recognition~\citep{turian2010word, huang2015bidirectional,akbik2018contextual, akbik2019pooled}, etc. Among these, \emph{CBOW}~\citep{mikolov2013efficient}, \emph{Skip-gram with Negative Sampling (SGNS)}~\citep{mikolov2013distributed}, \emph{GloVe} ~\citep{pennington2014glove} and \emph{FastText}~\citep{bojanowski2017enriching} are the most well-known ones and on which we experiment.

% what challenge
Despite the prevalence of word representations, only a minority of them can be directly applied to the raw informal text without preprocessing. Some specific word representations are proposed to such domain~\citep{tang2014learning, benton2016learning, dhingra2016tweet2vec, vosoughi2016tweet2vec}.
Conventional NLP systems usually do text normalization engineering on Twitter data at the preprocessing stage~\citep{jiang2011target, singh2016role, arora2019character}, by converting non-standard words to their standard forms. Nevertheless, the original meaning of raw sentences might have deviated to some extent~\citep{eisenstein2013bad}. The elimination of the spelling expression may lead to the scarcity of the writer's persona and behavioral characteristics information that the original demographic statistics can convey~\citep{saphra2016evaluating, benton2016learning}.

% my contribution
Therefore, we collected a variant spelling dataset for use of NLP research in the informal domain. Our key contributions are:
\setlist{nolistsep}
\begin{itemize}[noitemsep]
    \item to collect around 25k variant spelling pairs from \emph{UrbanDictionary};
    \item to employ weakly-supervised pattern-based bootstrapping and linear-chain CRF with self-training methods for extraction. Our results outperform the lexico-syntactic surface rule-based baseline method;
    \item to pretrain the word embeddings of \emph{CBOW}, \emph{SGNS}, \emph{GloVe} and \emph{FastText} on our cleaned English tweets and intrinsically evaluate them by measuring the cosine similarity using our variant spelling pairs;
    \item to evaluate a Twitter hashtag prediction downstream task with above 4 kinds of word embeddings and analyze the performance and correlation with the this intrinsic metric.
    \item to develop an online tool for searching informal word variant spelling.
\end{itemize}

\section{Related Work}
Word analogy task and word similarity task provide the staple benchmarks to evaluate the goodness of distributed word representations. 

\begin{table}[]
\centering
\begin{tabular}{@{}lll@{}}
\toprule
\textbf{relationship} & \multicolumn{2}{c}{\textbf{Word pair}}  \\ \midrule
Capital city                    & UK                & London               \\
Man-Woman                     & brother           & sister               \\
Pural nouns                   & mouse             & mice             \\
Present particle              & walk              & walking    \\ 
\bottomrule
\end{tabular}
\caption{Examples of syntactic and semantic analogical evaluation pairs in the word analogy task}
\label{tb:word-pair-examples}
\vskip -5mm
\end{table}

% intrinsic evaluation
\paragraph{Word analogy task}
To evaluate the quality of word representations, the word analogy task can serve as the intrinsic metric to measure the distance between analogical word pairs in the multi-dimensional word representation space~\cite{mikolov2013efficient, mikolov2013distributed, levy2015improving}. It requires the analogical word pair dataset that contains different semantic or syntactic relations such as (``UK'', ``London'') and (``King'', ``Queen''). MSR analogy dataset~\citep{mikolov2013linguistic} is proposed for this purpose. Four examples are shown in Table~\ref{tb:word-pair-examples}. The word analogy task measures the relation of ``a is to \~a as b is to \~b''. Taking the relation of capital city for instance, the word vector relation is like: vec(``London'') - vec(``UK'') +  vec(``France'') = vec(``Paris'') . 

\noindent Let $V$ denote the vocabulary of word representations, given the word pair (\emph{a}, \emph{\~a}) and relation (\emph{b}, \emph{\~b}), the word analogy task counts it correct only if the closest candidate word $\hat{a}$ is identical to the word \emph{\~a}:
\begin{align*} 
\hat{a} &= \argmax_{\hat{a} \in V \setminus \{\tilde{b}, b, a\}} \cos \left(\hat{a}, \tilde{b}-b+a \right) \\
         &= \argmax_{\hat{a}\in V \setminus \{\tilde{b}, b, a\}}\left( \cos(\hat{a},\tilde{b})-\cos(\hat{a},b)+\cos(\hat{a},a) \right)
\end{align*}

\paragraph{Word similarity task}
In addition, word similarity tasks deliver another intrinsic test metric on word embeddings \citep{pennington2014glove, levy2015improving}. Such datasets consist of \emph{WordSim353}~\citep{finkelstein2002placing}, \emph{MC}~\citep{miller1991contextual}, \emph{RG}~\citep{rubenstein1965contextual}, \emph{MEN} ~\citep{bruni2012distributional}, \emph{Mechanical Turk} ~\citep{radinsky2011word}, \emph{SCWS}~\citep{huang2012improving}, \emph{Rare Words}~\citep{luong2013better} and \emph{SimLex-999}~\citep{hill2015simlex}.
\noindent Similarly, given the synonym pair (\emph{a}, \emph{\~a}), we can get the nearest candidate word vector $\hat{a}$ as:
\begin{equation*}
     \hat{a} = \argmax_{\hat{a} \in V \setminus \{a\}} \cos \left(\hat{a}, a \right)
\end{equation*}{It considers the word representation capture the relationship beween \emph{a} and \emph{\~a} only when the word vector vec(\emph{\~a}) is closest to vec(\emph{a}) among all the vocabulary, i.e. $\hat{a} \equiv \tilde{a}$}.

% Naomi 2016 prev work a few pairs
The aforementioned evaluation word pair datasets are all aiming at formal domains, but there are few works of word pair datasets focusing on the informal domains.
\citet{saphra-lopez-2015-amrica} extracted a small size of variant spelling dataset from \emph{UrbanDictionary} using lexico-syntactic surface rule-based method with regular expression. Such manually summarized rule is devoid of diverse kinds of surface expressions, which might lead to some biases on the collected pairs despite its high precision.

Grounding on this, we employed semi-supervised approaches for the variant spelling dataset extraction and empirically illustrate its efficacy.

\section{Preliminaries}
\label{sec:pre}
We disentangle the informal spelling variant extraction task and clarify the definitions involved in this task.

\begin{table}
\centering
\begin{tabular}{ll}
\hline
\textbf{Item} & \textbf{Content} \\
\hline
informal word & \small {\color[HTML]{009901}\textbf{m8}} \\
definition & \small \makecell[l]{shorter way to say {\color[HTML]{329A9D}\textbf{mate}},especially \\over the internet}\\
example & \small "me and me m8s got pretty pissed." \\
author & \small International Bad Boy \\
update time & \small Sep 07, 2004 \\
upvotes & \small 1368 \\
downvotes & \small 195 \\ 
\hline
\end{tabular}
\caption{An example entry of the slang words in \emph{UrbanDictionary}. The correct spelling variant pair is ({\color[HTML]{009901}\textbf{m8}},{\color[HTML]{329A9D}\textbf{mate}}), where words in this pair has the same word lemma but with two variant spelling forms.}
\label{tb:urbandict-item}
\vskip -5mm
\end{table}

\paragraph{UrbanDictionary}  \emph{UrbanDictionary}\footnote{\url{https://www.urbandictionary.com/}} is a large-scale crowd-sourcing online slang dictionary, which records the newly emerging slang words, phrases and their meanings. It contains not only the informal words or phrases and their definitions but also definition tags, term editors and update time.  Table~\ref{tb:urbandict-item} presents the items that are contained in an example entry of \emph{UrbanDictionary}.
\noindent \emph{UrbanDictionary} holds the promise of collaborative NLP resources in the informal domain such as Twitter and social media forrums~\citep{nguyen2018emo}. It is also be utilized for Twitter text normalization task~\citep{beckley2015bekli} and explanation generation of unseen non-standard English language~\cite{ni2017learning}. Meanwhile, it could be quite independent of word representations trained on an informal text corpus such as tweets, assuring the fairness of the collected dataset. 

% % what is spelling variant
% \paragraph{Spelling variant} 

\paragraph{Spelling variant detection} As shown in Table~\ref{tb:word-pair-examples}, the word pair (``m8'',``mate'') can be regarded as spelling variants. For each spelling variant pair, the first instance belongs to informal words and the second one is a formal word. The spelling variant detection task is to extract the word pairs that possess the relations of spelling variants.
% task definition / aim

\noindent This task can be defined as: given a corpus of dictionary terms of size $n$, which consists of word entries $\mathbb{W} = \{w_0, w_1, w_2, ..., w_n\}$, corresponding definitions $\mathbb{D}=\{d_0, d_1,..., d_n\}$, and the expected relationship $\mathbb{R} := \textrm{spelling variants}$. For the $i^{th}$ word's definition with length $m$, we have $d_i = \{t_{i,0},\ t_{i,1},\ ...,\  t_{i,m}\} \ \textrm{where}\ i \in [0,n]$, $t_{i,m}$ represents the the $m$-th token in $i$-th word's definition. The purpose is to find the variant spelling tuple $y_i = (w_i, t_{i,v}) \ \textrm{where}\ v \in [0,m]$. Finally, we could get a large set of variant spelling tuples $\mathbb{T}=\{y_0,...,y_{z} \}$ where $|z|$ denotes the total count of extracted pairs.

% what is evaluation metric equation
\paragraph{Spelling variant similarity task} It is assumed that words with similar meanings would have closer distances in the distributed word vector space. The aforementioned word analogy task and word similarity task measure the cosine similarity between the given relation pairs. Likewise, taking the cosine distance between spelling variant pairs could allow for evaluating word representations pretrained on informal texts~\citep{saphra2016evaluating}. Taking the pair (``m8'', ``mate'') for instance, the pair score is denoted as 1 if vec(``mate'') ranks within top $k$ ($k \in \mathbb{R}^+$) closest word embeddings, otherwise 0. Afterward, the accuracy statistic takes the mean average precision (MAP) of all pairs as the overall performance. We term this task as \emph{spelling variant similarity task}.

\section{Approaches}
\label{sec:app}

\subsection{Baseline}
Target spelling variants in \emph{UrbanDictionary} are always accompanied by a certain kind of definition expression patterns, which can be used as the extraction rule. The direct way to spelling variant detection is summarizing lexico-syntactic surface rules and employ \emph{Regular Expression} (RE) techniques for extraction.

It is obvious that the matched definition expression patterns for spelling variants are capable of indicating the variant words, denoting as [Y]. For instance,
\begin{center}
\begin{tabular}{r}
% misspelling of [Y] \\
the variation of [Y] \\
another word for [Y] \\
another way of saying [Y] \\
% the most common form of [Y] \\
the incorrect spelling of [Y]
\end{tabular}
\end{center}

Accordingly we can manually create a large number of RE rules, as in Table~\ref{tb:RE}. Taking the pattern \verb|way of saying [Y]| for example, the RE pattern can be written as:

\begin{tabular}{l}
\verb-way of saying \"(?P<Spelling>[\w']+)\"- 
\end{tabular}

\subsection{Weakly-supervised pattern-based bootstrapping}

% \begin{figure}[thb]
\begin{figure}[!ht]
\vskip 0mm
\begin{center}
\includegraphics[width=1.2\columnwidth]{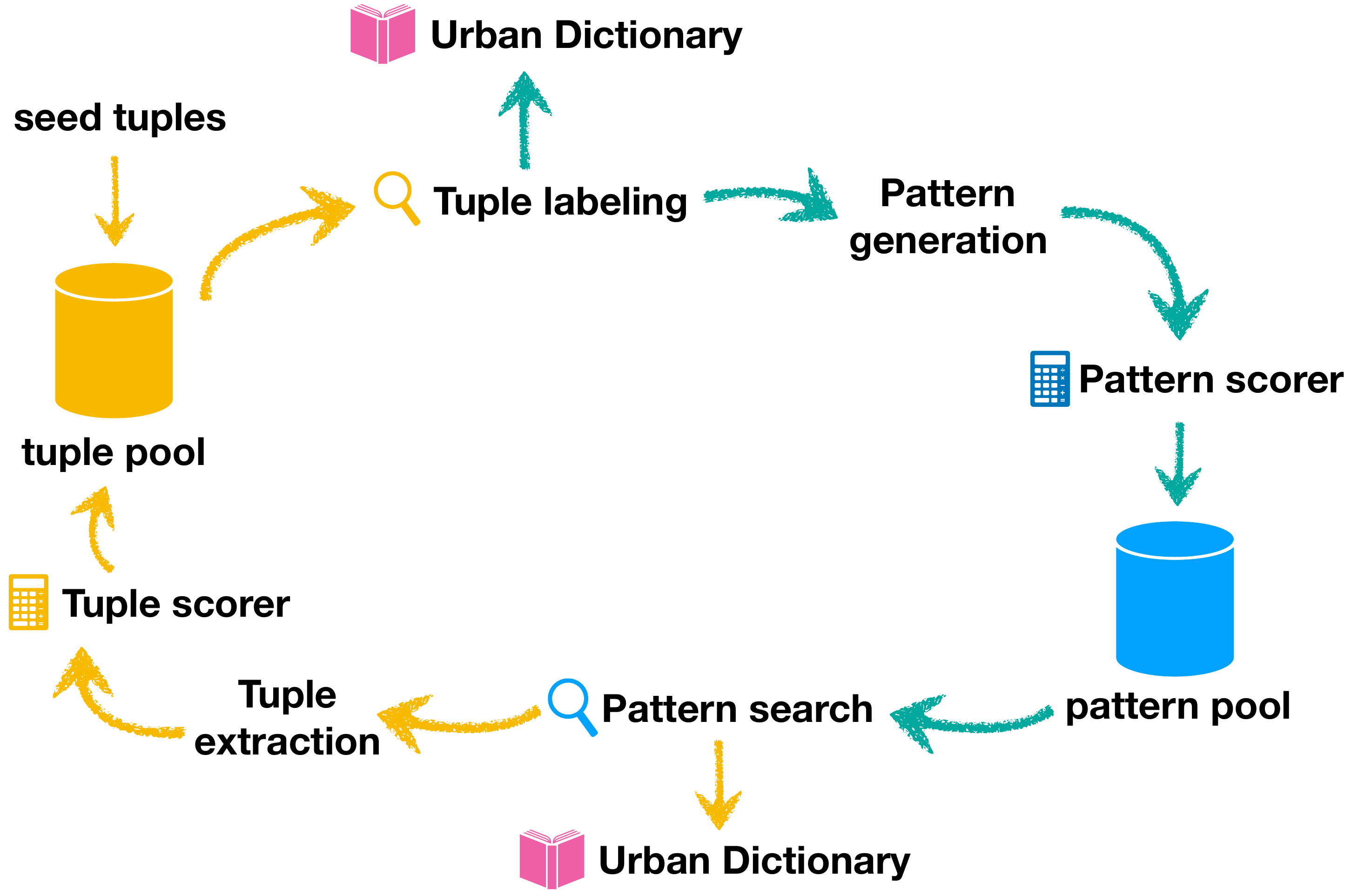}
\caption{The schematic diagram of weakly-supervised pattern-based bootstrapping method.}
\label{fig:bootstrap}
\end{center}
\vskip -5mm
\end{figure} 

Figure~\ref{fig:bootstrap} presents the semi-supervised pattern-based bootstrapping method to iteratively detect the target spelling variant pairs from unannotated corpus given a small set of seeds. A \emph{tuple pool} is used to label the unlabeled data and benefit for pattern generation, and it stores the initial given seeds and extracted candidate pairs with highly confidence in each iteration. Similarly, a \emph{pattern pool} incrementally appends matched relation patterns with fair certainty, on which the candidate tuples are matched accordingly. 

% \vskip 5mm
\begin{algorithm}[!h]
\small
% \SetAlgoLined
\SetKwInOut{Input}{Input}
% \SetKwInOut{Output}{Output}
% \KwResult{Write here the result }
\Input{unlabeled data $\mathbb{D}$ with size $|D|$, initial seeds \textbf{$S$}, confidence threshold of patterns and tuples: $\alpha \in [0.7,1)$ and $\beta \in [0.7,1)$, max iteration $M \in \mathbb{R}^+$
}
% \Output{ A set of variant spelling pairs}
 \textbf{Initialize}: tuple pool $T$ with $S$, empty pattern pool $P$\;
 \For{i = $\{1,2,3,\cdots, M\}$}{
  tuple labeling $\gets$ searching the occurrence of pair instances in $S$\;
  patterns $\gets$ use occurrence context to generate patterns\;
  
  \For{pat \textbf{in} patterns}{
  $\textrm{pat.score} \gets \textrm{scorePatterns(pat)}$\;
  \eIf{$\textrm{pat.score} > \alpha$}{
       $P \gets P + \textrm{pat}$\;
   }{
       del pat\;
  }
  }
  tuples $\gets$ P.matchCandidateTuple($\mathbb{D}$))\;
   \For{tup \textbf{in} tuples}{
   tup.score  $\gets$ scoreTuple(tup)\;
   \eIf{tup.score $> \beta$}{
       T $\gets$ T + tup\;
   }{
       del tup\;
    }
   }
 }
 \textbf{return} T\;
\caption{Weakly-supervised pattern-based bootstrapping algorithm}
\label{alg:bt}
\end{algorithm}

\subsubsection{Bootstrapping process}
Algorithm~\ref{alg:bt} illustrates the procedure of our bootstrapping algorithm. Firstly, initialize the tuple pool with a small size of $m$ seeds and label the occurrence of instances of seeds in the tuple pool. Afterward, generate candidate patterns grounded on the context of the previous occurrence and pass them to a pattern scorer. The filtered patterns are appended into the pattern pool and are used to search new tuples in turn. These candidate tuples are merged into the tuple pool after passing through a candidate tuple scorer. Finally, go to the first step until reaching the maximum iteration. 

\paragraph{Pattern generation}
When generating candidate patterns, we use the contextual words with a fixed context size $w$ on both sides. Given the definition ``Scottish way of saying \textbf{yes}'' and spelling variant tuple (``aye'', ``yes'') for example, the generated candidate pattern will be \verb-way of saying<\w>- when the window size $w=3$. 
% Besides, the pos tags of context can also be considered in the patterns.

\begin{table}[thb]
\begin{tabular}{ll}
\hline
definition    & generated pattern   \\ \hline
scottish way of saying \small{\color[HTML]{009901}\textbf{yes}} & way of saying {\small \verb-<\w>-} \\
another word for \small{\color[HTML]{009901}\textbf{weed}}  & another word for {\small \verb-<\w>-} \\
another way to say \small{\color[HTML]{009901}\textbf{your}} & way to say {\small \verb-<\w>-} \\ \hline
\end{tabular}
\caption{Three examples of generating surface patterns with context window size $w$=3. The tuples are (``aye'',``yes''), (``dank'',``weed'') and (``ur'',``your'')}. The formal words of tuples are in green.
\label{tab:pat_egs}
\vskip -5mm
\end{table}

\paragraph{Pattern scorer}
A well-defined pattern scorer plays a crucial role in bootstrapping methods because the candidate patterns generated from the previous step could face two main problems\cite{gupta2014improved}:
\begin{itemize}[noitemsep]
    \item over-confidently assign the badly-behaved pattern with a high score;
    \item conservatively treat the well-performed pattern with low confidence.
\end{itemize}{The ideal patterns are expected to reach a balance between accuracy and coverage. }
The candidate patterns in the preceding iteration could affect the following steps to a great extent. Virulent patterns could match pernicious seeds and such seeds could be regarded as reliable to generate new patterns. This effect could propagate iteration by iteration and finally destroy the system performance.

To avoid those problems and better evaluate the confidence of patterns. \emph{RlogF} scoring metric is employed in our system\citep{riloff1996automatically}. The score of $i$-th pattern can be defined as:
\begin{equation*}
    \textrm{score}(\textrm{pattern}_i) = \frac{F_i}{N_i} \log_2(F_i)
\end{equation*}{where $F_i$ denotes the count of unique variant positive entities in the tuple pool that $i$-pattern match, $N_i$ represents the total count that $i$-th pattern can extract.}
It can be seen that this scoring metric attends to both accuracy and coverage. The factor $\frac{F_i}{N_i}$ is high when variant pairs extracted by the $i$-th pattern have good coverage and correlation with reliable tuples in the tuple pool\cite{carlson2010active}. Meanwhile, the factor $F_i$ indicates the coverage that $i$-th candidate pattern can achieve. To make it conservative, we set a high predefined threshold for each pattern score to only retain the most reliable patterns at each iteration.

\paragraph{Tuple scorer}
The quality of extracted tuples can also have a huge influence on the final results. A small set of the wrong tuple may lead to \emph{semantic drift}, the malignant cycle in the bootstrapping iteration. In other words,  wrongly extracted tuples could lead to a couple of problematic patterns, and in turn, the noisome patterns could match amounts of destructive candidate pairs.

Hence we calculate the confidence for $i$-th candidate in the tuple pool as:
\begin{equation*}
    \text{score}(\text{tuple}_i) = \frac{\sum_{j=1}^{P_i} \log_2 (F_j+1)}{P_i}
\end{equation*}{where $P_i$ is the count that patterns are able to extract tuple $i$ and $F_i$ signals the unique number of tuples in the tuple pool that can be extracted by pattern $j$.}

Intuitively, candidate tuples that can be matched by multiple patterns are regarded as more reliable than those only matchde by one pattern~\cite{carlson2010active}. If tuple $i$ can be extracted by a large number of patterns, the tuple is comparably more confident than others.

It can be seen that the aforementioned tuple scorer only considers the relevance between positive tuples and reliable patterns, but ignore the occurrence count of candidate pairs. We assume that if a candidate pair occurs multiple times, it is more likely to be a target tuple. Hence we define a variant called \emph{RlogF with tuple count}:
\begin{equation*}
    \text{score}(\text{tuple}_i) = \frac{\sum_{j=1}^{P_i} \log_2 (F_J+1)}{P_i} \log_2|\text{tuple}_i|
\end{equation*}{where $|\cdot|$ is the occurrence count in the training data.}

Finally, we only maintain the top $N$ most confident tuples and remove others which are less reliable.

\paragraph{Target word constraint}
Some extraction tasks like name entity recognition (NER) could utilize existing toolkit for Part-of-Speech(POS) tagging ahead of time so that the bootstrapping system could easily find the chunk boundary of each candidate. But the un-pre-tagged data is quite hard to detect the correct chunk boundaries. Taking the definition of ``\emph{it's a way of saying \text{someone is really loud}}'' for example, what the surface pattern \verb-a way of saying- can match is a wrongly extracted word [\emph{someone}] instead of a clause [\emph{someone is really loud}] in this case.

Inspired by \emph{SPIED}~\citep{gupta2014spied} that adds the POS tag constraints on target extracted entities, we adopt the stopword constraints to filter out the wrongly extracted entities: if the extracted entity is a stopword such as ``the'' and ``something'', the normalized Levenshtein distance between the instances in extracted candidate tuples is used to measure the similarity of the formal and informal words. Normalized Levenshtein distance is dividing the Levenshtein distance by the overall length of the word pairs. Only word pairs whose score values are lower than a threshold $\tau$ could be retained.

\subsection{Self-training based linear chain CRF tagging}
The linear chain CRF is often applied in the sequence tagging in a supervised way. With self-training strategy, we iteratively train the supervised CRF system on the unlabeled data.

As illustrated in Algorithm~\ref{alg:self-train}, we firstly use the linear chain CRF model trained on a small size of labeled data $X$ to predict a large number of unlabeled data $U$, and supplement the most confident samples into initial golden labeled data. Then the next iteration will be run using the updated $X$ until satisfying the stop criteria.

\begin{algorithm}
% \SetAlgoLined
\small
\SetKwInOut{Input}{Given}
\Input{Small amount of labeled word sequence $X$ with positive and negative tags `I' and `O', a large number of unlabeled data $U$, maximum iteration $N \in \mathbb{R}^+$, confidence threshold $\tau \in [0.7, 1]$}
\SetKwInOut{Output}{Output}
\Output{The most confident label sequence $Y$ with the highest conditional probability $P(Y|X)$}
    $\#$ load gold labeled data for the first iteration;
    X = \text{load\_labeled\_data}() ;
\For{$i \gets \{1, 2,\cdots, N\}$ }{
    $\mathcal{F}_x = \text{feature\_generation}(X)$;
    $crf = \text{CRF}(\mathcal{F}_x)$;
    $ \text{SilverData} = \{\}$;
    \For{\text{each sample} u \textbf{in} U}{
      \textrm{u.tag}$ \gets crf.predict(u)$;  
      \For{\text{word} \textbf{in} u}{
          \eIf{\text{word.tag} == `I' \&\& \text{word.tag}$ > \tau$}{
           \text{word.tag}$\gets$ `I';
           \textrm{SilverData}$\gets \text{SilverData} + u$;
           $U \gets U - u$;
           \textbf{continue};
        }{\text{word.tag}$\gets$ `O';}
        }
      $X \gets X + \textrm{SilverData}$;
    }
 }
 
 \caption{Self-training based CRF system}
 \label{alg:self-train}
\end{algorithm}

\paragraph{Feature engineering}
Feature engineering is extremely important for most statistical machine learning systems. Unlike the aforementioned methods, we use both shallow features
and deep-parsed text features, such as dependency parsing and
POS tag information.

% \begin{figure}[thb]
\begin{figure}[!ht]
\vskip 0mm
\begin{center}
\includegraphics[width=\columnwidth]{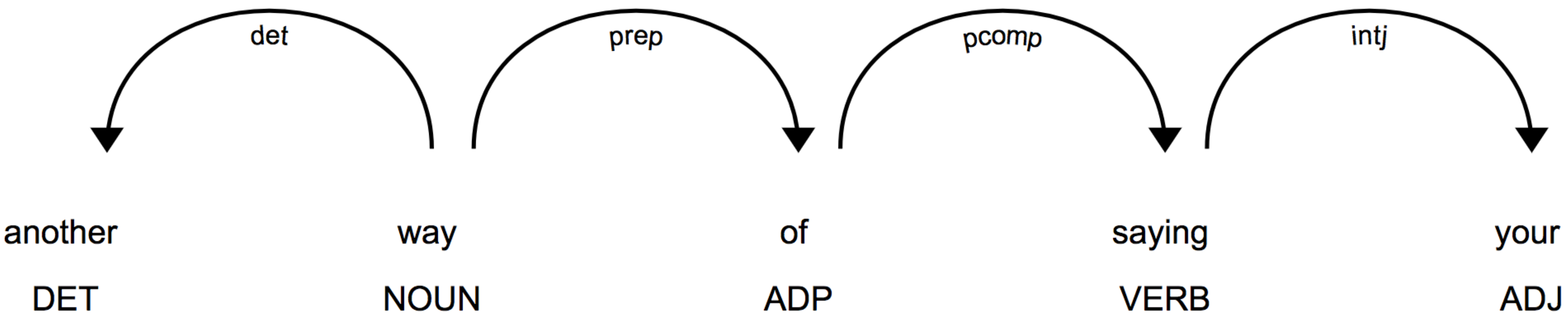}
\caption{The dependency parsing of word \textit{ur}'s definition: ``another way of saying your''.}
\label{fig:dep}
\end{center}
\vskip -5mm
\end{figure} 

\noindent For each target word, the features including:
\begin{itemize}
    \item word lowercase
    \item word lemma
    \item whether it is digit
    \item whether it is title
    \item POS tag\footnote{including both shallow POS tags (e.g. VERB) and detailed POS tag features (e.g. VBG), denoting ``$pos\_$'' and ``$tag\_$'' in the following sections. }
    \item syntactic dependency\footnote{\url{https://nlp.stanford.edu/software/dependencies manual.pdf}}
    \item the word lemma and POS tag of its head word in the dependency path tree
\end{itemize}{}
\noindent For the context of the target words, we include:
\begin{itemize}
   \item word lowercase
   \item word lemma
   \item whether it is a digit
   \item whether it is a title
   \item POS tag
   \item the word lemma and POS tag of its head word in dependency path tree
   \item The end of sentence tag \emph{EOS} or the beginning of sentence \emph{BOS}
\end{itemize}{Table~\ref{tb:feat_eg} illustrates the features extracted for the definition in Figure~\ref{fig:dep} with context window size 1.}

\section{Experiments}
\label{sec:exp}
The spelling variant pair detection is done with the following methods. Finally, we can get a large list of spelling variant pairs.

% baseline
\subsection{Baseline} 
In the baseline, the RE patterns are directly employed used for variant spelling detection. The matched variant entity counts vary from 9 to 951 among our summarized surface patterns as shown in Table~\ref{tb:RE}. We randomly permute and sample 200 pairs from them and manually calculated accuracy is approximately 80\%. The baseline approach has high accuracy but low coverage. The extracted patterns are simple and monotonic due to the limited expensive hand-crafted rules. It is laborious and impractical to detect the possible rules for all corpus because rules usually require specific domain knowledge and do not have flexible scalability.

\subsection{Bootstrapping methods}

\begin{figure}[!ht]
% \vskip 5mm
\begin{center}
\centerline{\includegraphics[width=\columnwidth]{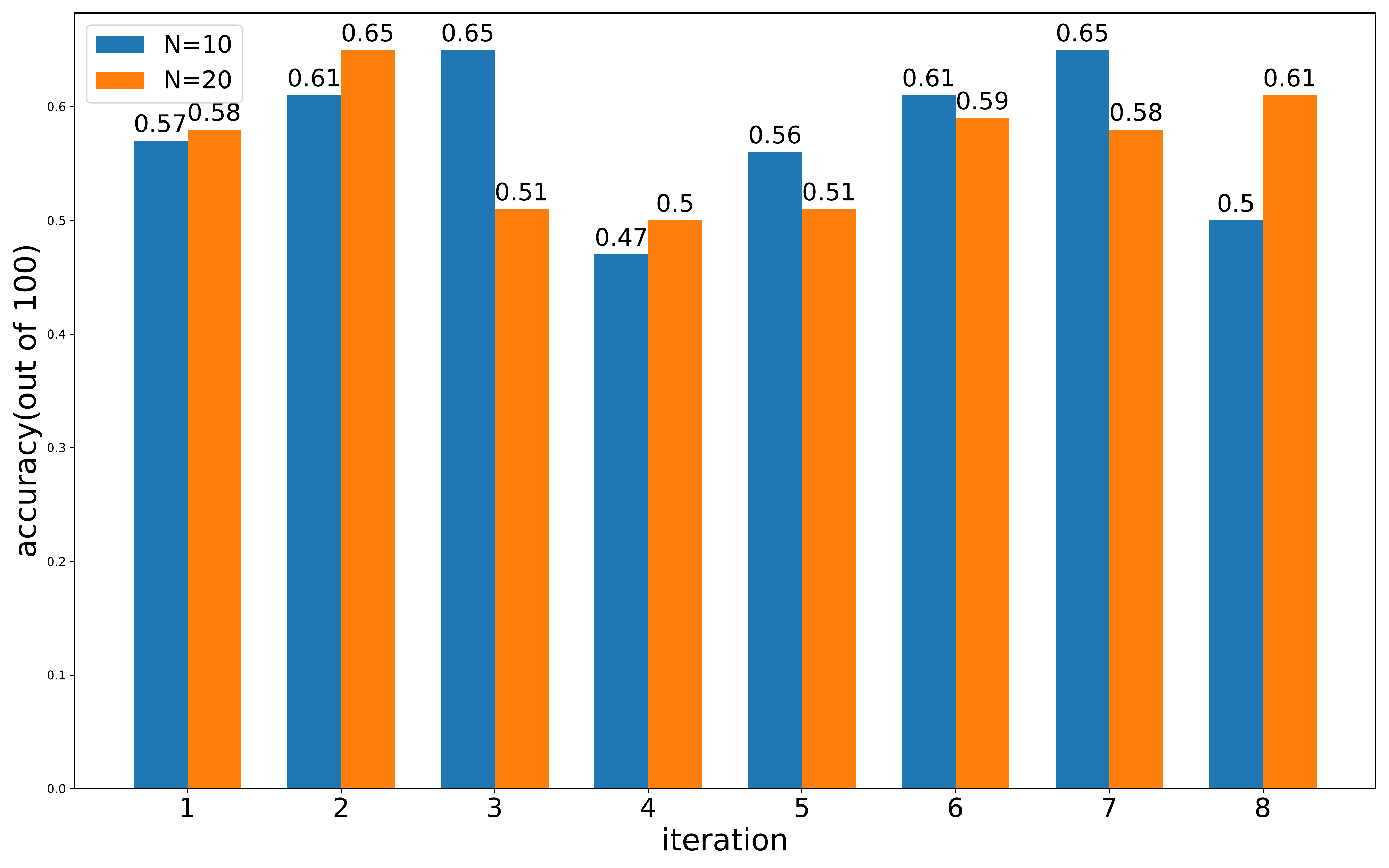}}
\caption{The accuracy of basic boostrapping methods with tuple scorer selection size $N \in \{10,20\}$.}
\label{fig:bt_size}
\end{center}
\vskip -5mm
\end{figure} 

We set the initial seed size as 5, iteration count 8, selected top-ranked size of pattern scorer as 10, the threshold of normalized Levenshtein distance $\tau=0.5$, the pattern context size $w=3$ and the selected ranked tuple size $N \in \{10,20\}$. We manually estimate the accuracy from sampled 100 extracted tuples for each experiment.

% \begin{figure}[!h]

As shown in Figure~\ref{fig:bt_size}, the bootstrapping systems selected the top $N$ tuples after tuple matching. It can be seen that systems with $N=20$ outperforms the counterpart with $N=10$ at the initial two iterations and the trend becomes opposite thereafter. It may be because that the detected tuple count at first is fewer than the limited size $N$, but systems with larger filter size could include less confident tuples. Such noise candidates could lead to a vicious cycle in the bootstrapping process and deviate away from the correct results. Systems with $N$=10 and 20 can extract 28,397 and 29,643 spelling variant pairs at the end of the 8-th iteration.

\begin{figure}[thb]
% \vskip 5mm
\begin{center}
\centerline{\includegraphics[width=\columnwidth]{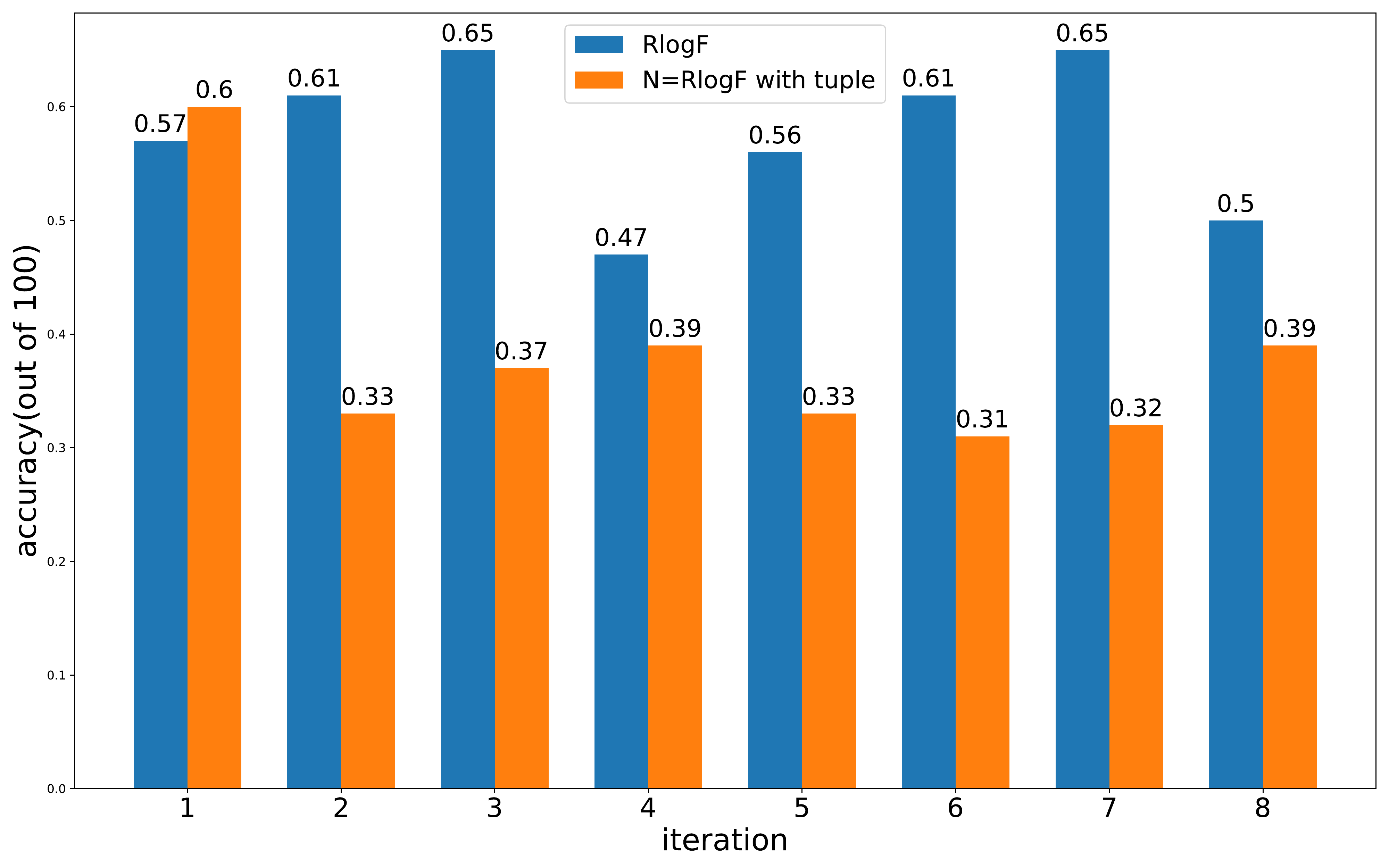}}
\caption{the accuracy of bootstrapping systems with tuple scorer of \emph{RlogF} and \emph{RlogF with tuple count}. The top 10 ranked tuples are retained at each iteration.}
\label{fig:bt_RlogF}
\end{center}
\vskip -5mm
\end{figure} 

Table~\ref{fig:bt_RlogF} presents that the bootstrapping systems degrade when we consider matched tuple count based on \emph{RlogF} tuple scorer. This may be due to the semantic drift of the iterative process. We find that adding the tuple count factor results in more noisy patterns such as \emph{a typo of, of the verb, to type out}. This might be because the factor counts too much and should be scaled or regularized.

\begin{figure}[!ht]
% \vskip 5mm
\begin{center}
\centerline{\includegraphics[width=\columnwidth]{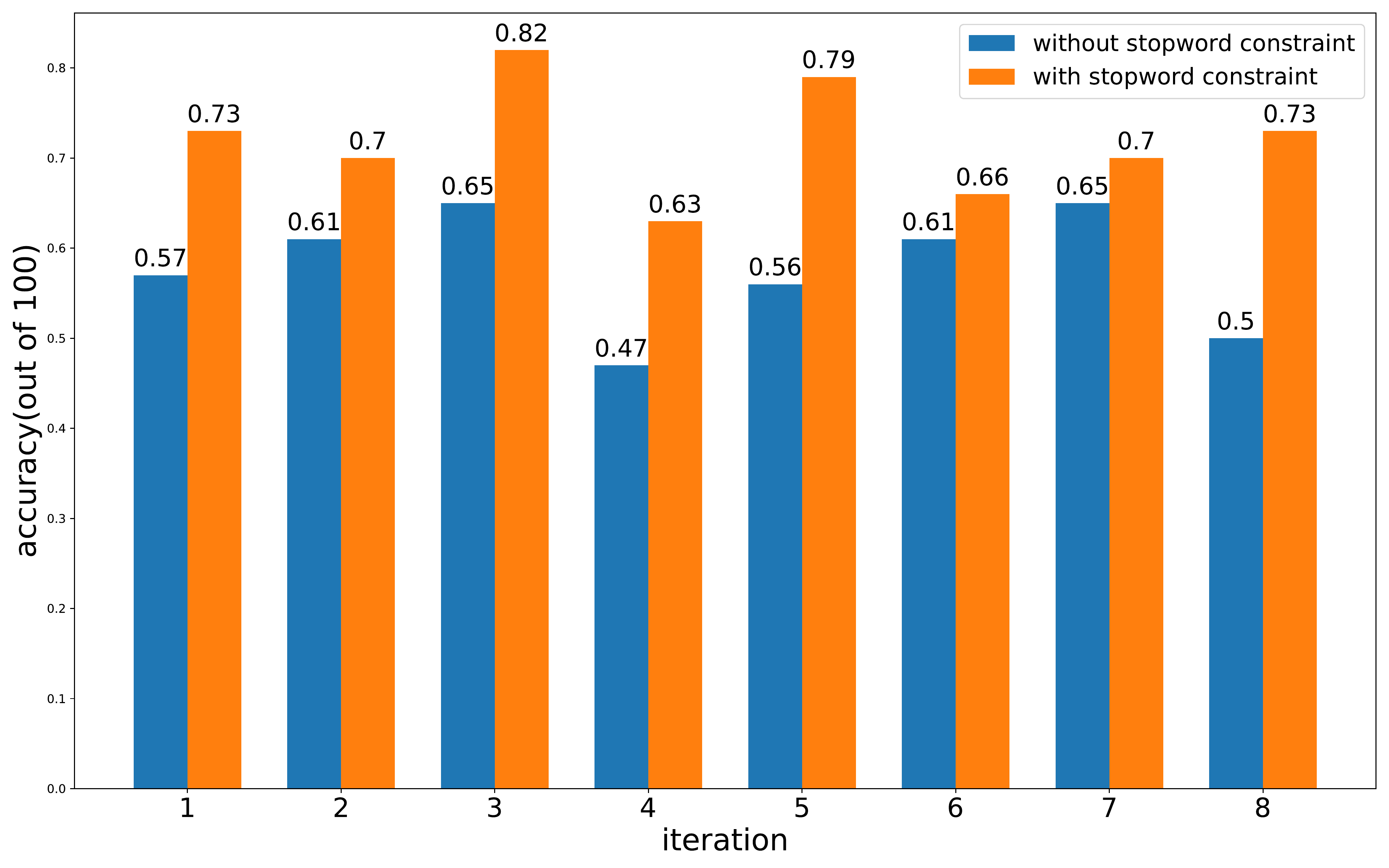}}
\caption{The accuracy of boostrapping systems with and without stopword constraint.}
\label{fig:bt_stopword}
\end{center}
\vskip -5mm
\end{figure} 

We can observe in Figure~\ref{fig:bt_stopword} that the accuracy increases and keeps stable with the increase of iteration count after adding stopword constraint. The average and maximum accuracy of the detection tasks is 72\% and 82\% respectively. This trick makes the bootstrapping systems more conservative and guarantees its high accuracy.

\subsection{Self-training}
To suffice the training requirement of self-training, we firstly manually labeled 2,000 samples that contain positive variant entities and 1,000 samples without variant words. We employ \textbf{IO} tagging method, where \textbf{I} represents ``inside'' the variant entity boundary and \textbf{O} denotes ``outside'' the target variants. 

We set the context size $w \in \{3,4\}$, confidence threshold $\tau \in \{0.8, 0.9\}$, maximum iteration number as 5. Similarly, we manually check the correct tuples from sampled 100 instances. 

\begin{figure}[!ht]
% \vskip 5mm
\begin{center}
\centerline{\includegraphics[width=\columnwidth]{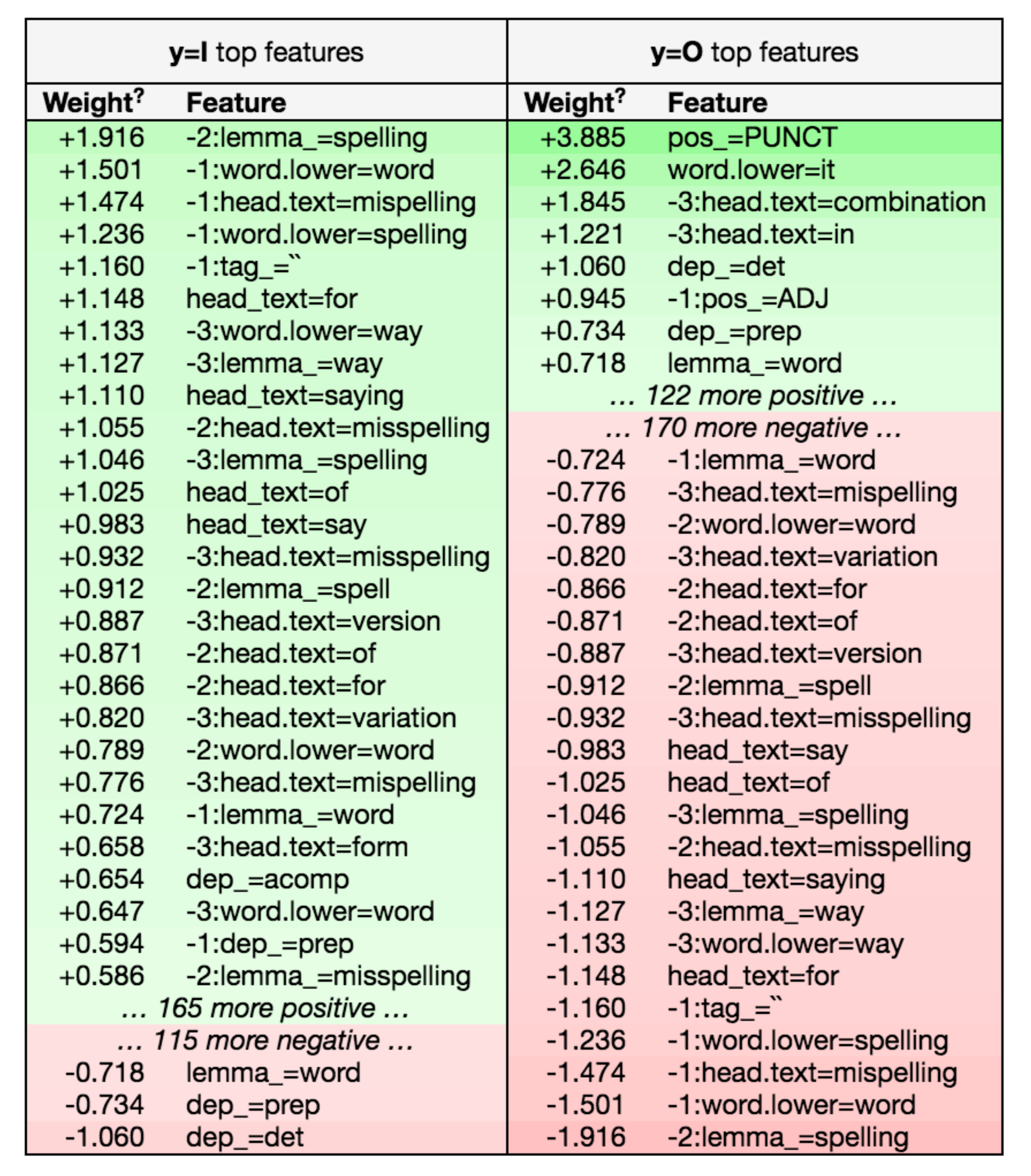}}
\caption{The contribution of selected features in CRF model, where
`dep\_', `pos\_', `tag\_' represent the dependency tag, shallow POS tag and detailed POS tag respectively. The contribution in green and in red means the positive and negative contribution respectively. The color saturation denotes the degree of the corresponding effect.}
\label{fig:crf_feat}
\end{center}
\vskip -5mm
\end{figure} 

\paragraph{Random search}
We employ \emph{L-BFGS} training algorithm with Elastic Net regularization to train the CRF model with the random search for hyperparameter tuning techniques. We randomly sample the hyper-parameter 50times and use 3-fold cross-validation to tune the initial CRF model on the hand-labeled small amount of gold-labeled data. The CRF achieves the best with L1 value 2.35 and L2 value 0.08.

\begin{figure}[thb]
% \vskip 5mm
\begin{center}
\centerline{\includegraphics[width=\columnwidth]{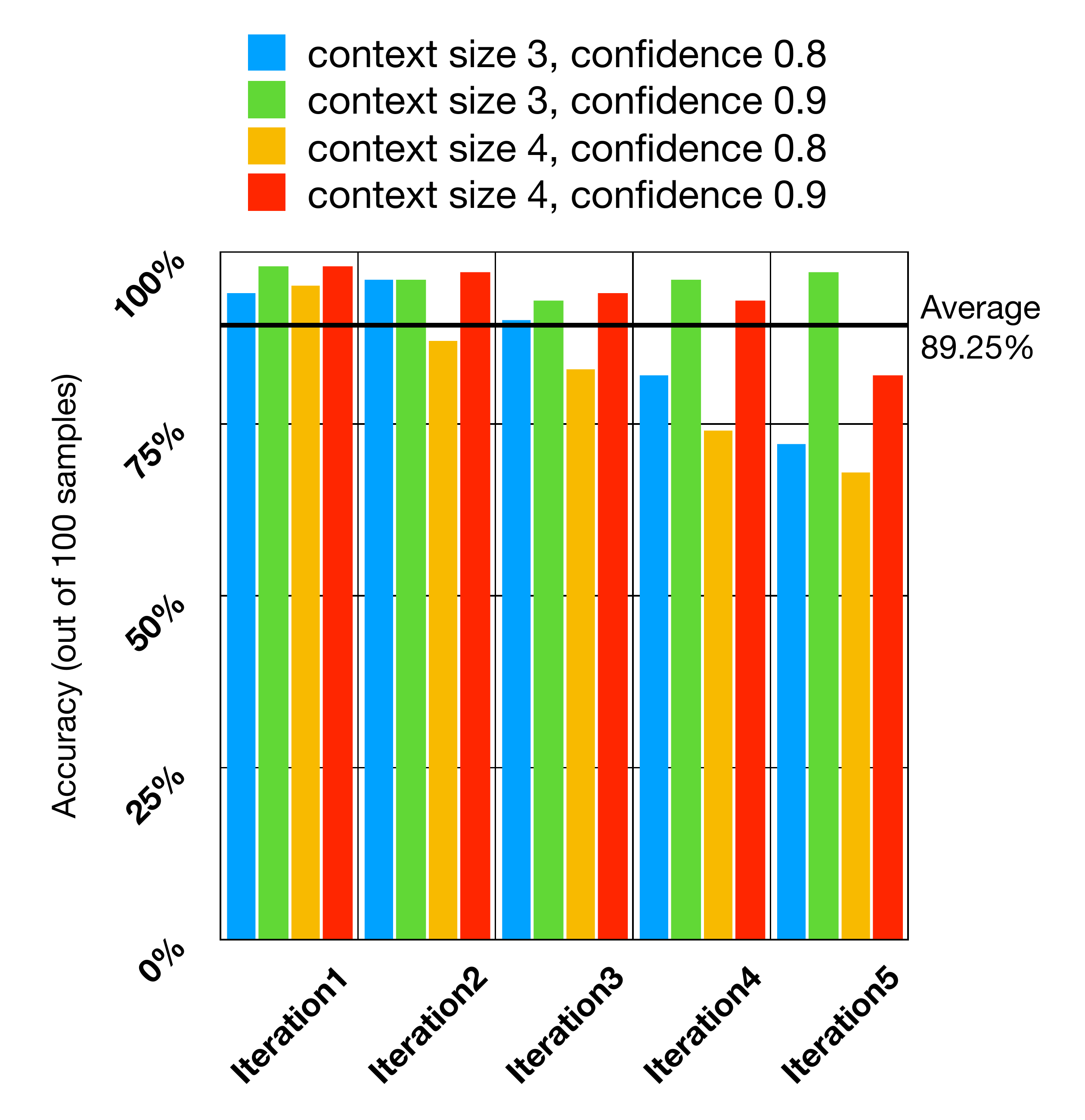}}
\caption{The accuracy of CRF models with context size of 3,4 and confidence threshold of 0.8,0.9.}
\label{fig:crf_acc}
\end{center}
\vskip -5mm
\end{figure} 

\paragraph{Feature contribution}
We generated features with a fixed context size 3, similar to examples shown in Figure~\ref{tb:feat_eg}. We plot the contribution for selected features in Figure~\ref{fig:crf_feat}. Suppose the current word position is $i$, we can see that when the $\{i-2\}$-th word is ``spelling'', the current word is more reliable to be the target entity. This supports the surface rule ``another spelling of''. Besides, when the POS tag of the current word is punctuation or the word's lowercase is the word ``it'', it is less confident to be the target. Such ranking has a great match on the intuition of our domain-dependent rules.

Afterwards, 5 iterations are run for each model setting, as in Figure~\ref{fig:crf_acc} and~\ref{fig:crf_num}. The model with the context size 3 and confidence threshold 0.9 achieves the best accuracy at above 90\%. In contrast, the model with context size 4 and confidence threshold 0.8 identifies the most tuple numbers but presents a decreasing low accuracy during all the iterations. Thus it is necessary to reach a balance between the extracted number (coverage) and precision. Finally, we choose context size 3 and confidence threshold 0.9 as the optimal setting, where the accuracy is 0.97 and the extracted count is 26,698 at iteration 5. It can be seen that self-training approaches outperform the previous two methods in terms of accuracy.

We also developed an online tool for the purpose of searching for spelling variants shown as figure~\ref{fig:web-search}, ~\ref{fig:web-result}.

\subsection{Analysis}
We train the aforementioned four kinds of embeddings including \emph{CBOW}, \emph{SGNS}, \emph{GloVe} and \emph{FastText} on self-cleaned 2.35GB English Tweets from scraped 260GB multi-lingual tweets. The preprocessing consists of tokenization, lowercase and removing non-English words, but without text normalization.

These embeddings are trained with minimum vocabulary occurrence of 200, the context window size of 5/10/15 and embedding size of 100/200.

\paragraph{Spelling variant similarity}
We evaluate the pre-trained word representations on the spelling variant similarity task described in section~\ref{sec:pre}. Here we use English Simple wiki and Wiki data to generate the formal word vocabulary and filter out the word tuples whose second instances are not formal words. The MAP values of cosine similarity with top $k$ of 1/20/50/100 are shown in Table~\ref{tb:res-simp},~\ref{tb:res-en}.

\paragraph{Twitter hashtag prediction}
We also experimented on a Twitter hashtag prediction task with conventional TextCNN classification models to compare the correlation between the spelling variant similarity task and the performance on the hashtag prediction task. The results is in Table~\ref{tb:pred_res}.

\begin{figure}[!ht]
% \vskip 5mm
\begin{center}
\centerline{\includegraphics[width=\columnwidth]{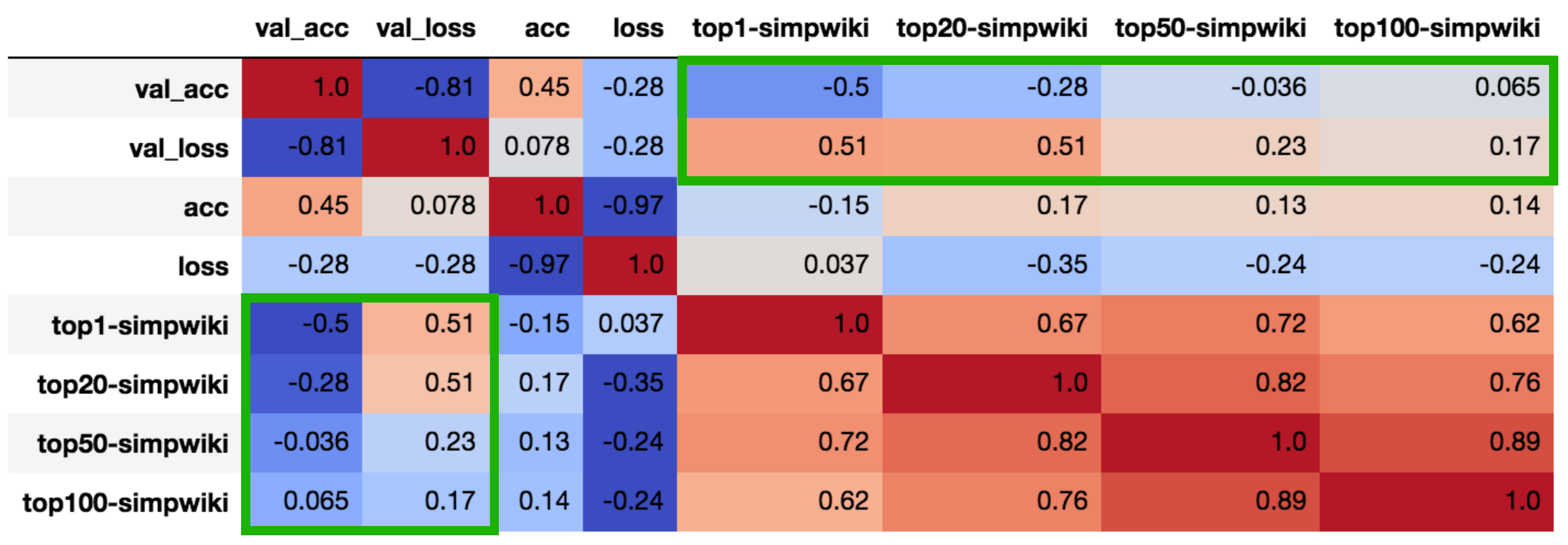}}
\caption{  \textbf{Pearson correlation} between the extrinsic model performance and the intrinsic metric using \textbf{simple wikipedia} as the formal vocabulary.
}
\label{fig:pearson-simp}
\end{center}
\vskip -5mm
\end{figure}

\paragraph{Comparison}
The \emph{Pearson correlation} between the previous intrinsic and extrinsic metrics are in Figure~\ref{fig:pearson-simp},~\ref{fig:pearson-en}. Their performance has a certain correlation in terms of its best loss. As for measuring the similarity of the top $k$ closest words, the correlation increases with the decrease of the value $k$. When we compute the top 1 similarity, the correlation could be relatively high. 

The scatter plot of best loss and accuracy on dev set are in Figure~\ref{fig:scatter-simp-loss},~\ref{fig:scatter-simp-acc},~\ref{fig:scatter-en-loss} and~\ref{fig:scatter-en-acc}. Figure~\ref{fig:scatter-simp-loss} and~\ref{fig:scatter-en-loss} both reflect the positive correlation between the performance of such two tasks.

It can be concluded that the \emph{spelling variant similarity task} perform relatively correlated with the Twitter hashtag prediction downstream tasks. Such evidence promotes the odds of removing the text normalization in the NLP pipelines and directly use embeddings trained on the informal-domain text.

\section{Conclusion}
We have extracted a variant spelling tuple dataset of approximately 25K tuples that could achieve a roughly above 90\% accuracy. We empirically prove that the text normalization may be removed when handling NLP tasks in the informal domain. Such a spelling variant dataset can also be used in a large number of NLP systems of the informal domain.

% \section*{Acknowledgments}

\bibliography{anthology,acl2020}
\bibliographystyle{acl_natbib}

\clearpage
\thispagestyle{plain}
\appendix
\section{Appendices}
\label{sec:appendix}

\begin{landscape}
\begin{table}[]
\centering
\begin{tabular}{|l|c|l|}
\hline
Surface regular expression rule & matched count  & Example tuple \\ \hline
\begin{minipage}{4in}
\begin{spverbatim}
spelling[^\.,]{0,3}?( of| for| to|:| the word| include|)[^\.,]{0,5}?\"(?P<Spelling>
\w+)\"
\end{spverbatim}
\end{minipage}
& 951 & (kewl, cool), (crao, crap) \\ \hline
\begin{minipage}{4in}
\begin{spverbatim}
spelling[^\.,]{0,3}?( of| for| to|:| the word| include|)[^\.,]{0,5}?'(?P<Spelling>
\w+)'
\end{spverbatim}
\end{minipage} & 306 & (dentisit,dentist), (yuo, you) \\ \hline
 \begin{minipage}{4in}
\begin{spverbatim}
^meaning \"(?P<Spelling>[\w']+)\"
\end{spverbatim}
\end{minipage}& 38 & (bewtuh, better), (nair, no) \\ \hline
\begin{minipage}{4in}
\begin{spverbatim}
^meaning '(?P<Spelling>\w+)'
\end{spverbatim}
\end{minipage} & 9 & (oned, owned), (fidoosh, finished) \\ \hline
\begin{minipage}{4in}
\begin{spverbatim}
way of saying \"(?P<Spelling>[\w']+)\"
\end{spverbatim}
\end{minipage} & 843 & (Ogay, Okay), (gorl, girl) \\ \hline
\begin{minipage}{4in}
\begin{spverbatim}
way of saying '(?P<Spelling>\w+)'
\end{spverbatim}
\end{minipage} & 234 & (heauge, huge) , (Ochea, OK) \\ \hline
\begin{minipage}{4in}
\begin{spverbatim}
form of [^\.,]{0,3}?\"(?P<Spelling>[\w']+)\"
\end{spverbatim}
\end{minipage} & 527 & (oof, oops), (F8, faight) \\ \hline
\begin{minipage}{4in}
\begin{spverbatim}
form of '(?P<Spelling>\w+)'
\end{spverbatim}
\end{minipage} & 151 & (bab, baby), (gr8, great) \\ \hline
\begin{minipage}{4in}
\begin{spverbatim}
^short for \"(?P<Spelling>[\w']+)\"
\end{spverbatim}
\end{minipage} & 104 & (fend, defend), (fied, satisfied) \\ \hline
\begin{minipage}{4in}
\begin{spverbatim}
^short for '(?P<Spelling>\w+)'
\end{spverbatim}
\end{minipage} & 19 & (inet, internet), (hols, holidays) \\ \hline
\end{tabular}
\caption{The manually created regular expressions to match lexico-syntactic surface patterns.}
\label{tb:RE}
\end{table}
\end{landscape}

\begin{table*}[thb]
\centering
\begin{tabular}{lll}
\hline
current word features & previous -1 context features & following +1 context features \\ \hline
word.lower=your       & -1:word.lower=saying         & +1:EOS                        \\
word.istitle=False    & -1:word.istitle=False        &                               \\
word.isdigit=False    & -1:word.isdigit=False        &                               \\
pos\_=ADJ             & -1:pos\_=VERB                &                               \\
tag\_=PRP             & -1:tag\_=VBG                 &                               \\
dep\_=intj            & -1:lemma\_=say               &                               \\
lemma\_=your        & -1:dep\_=pcomp               &                               \\
head\_text=saying     & -1:head.text=of              &                               \\
head\_pos=VERB        & -1:head.pos\_=ADP            &                               \\
head\_tag=VBG         &                              &                               \\ \hline
\end{tabular}
\caption{The extracted features (window size: 1) of the word \textit{your} in the definition example in figure~\ref{fig:dep}.}
\label{tb:feat_eg}
\end{table*}

\begin{figure*}[!h]
% \vskip 5mm
\begin{center}
\centerline{\includegraphics[width=1.4\columnwidth]{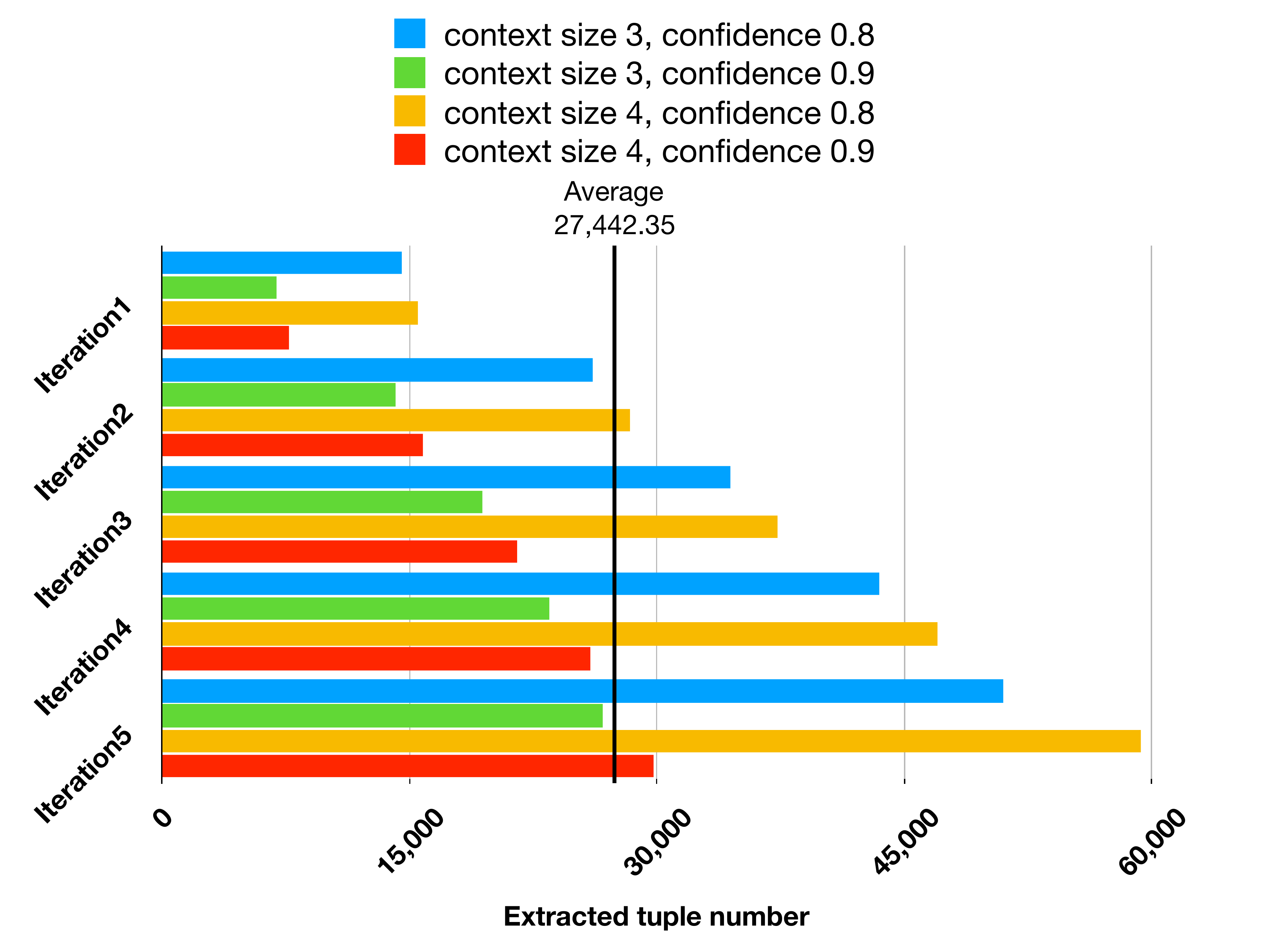}}
\caption{The extracted count for the self-training based CRF model, with the context size of 3,4 and the confidence threshold 0.8,0.9.}
\label{fig:crf_num}
\end{center}
\vskip -5mm
\end{figure*}

\begin{figure*}[thb]
\vskip -2mm
\begin{center}
\centerline{\includegraphics[width=\columnwidth]{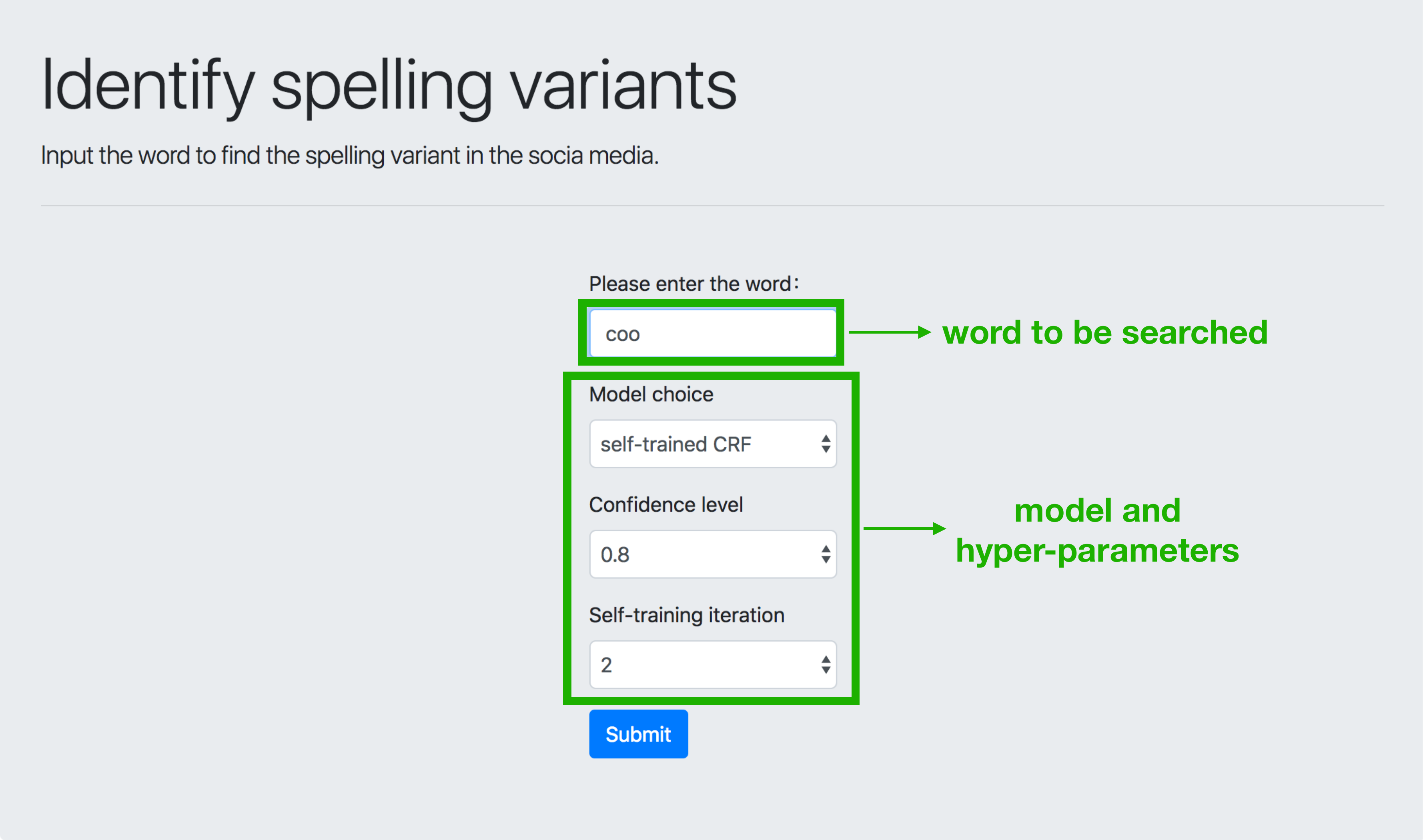}}
\caption{The query interface of variant spelling. The results are scraped and extracted from real-time Urban Dictionary corpus. After searching for a specific informal word, it will return all the Urban Dictionary definitions, display and highlight all the discovered variant spelling entities.}
\label{fig:web-search}
\end{center}
\vskip -5mm
\end{figure*}

\begin{figure*}[thb]
\vskip -5mm
\begin{center}
\centerline{\includegraphics[width=\columnwidth]{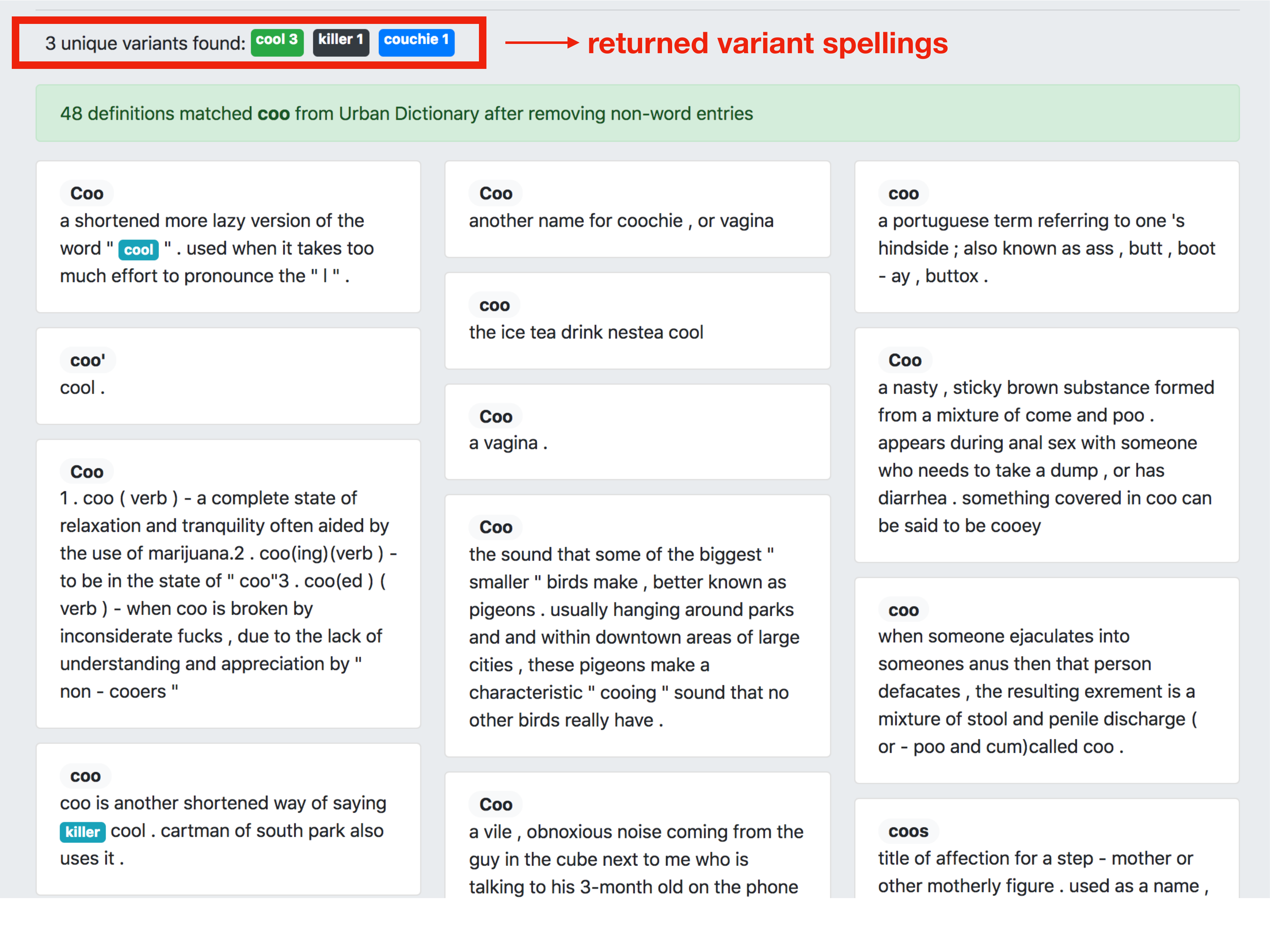}}
\caption{Results of found variant spelling and definitions.}
\label{fig:web-result}
\end{center}
\vskip -5mm
\end{figure*}

\begin{table*}[!h]
\resizebox{\textwidth}{!}{\begin{tabular}{c|cc|ccccc}
\hline
\multicolumn{3}{c|}{embedding} & \multirow{2}{*}{\begin{tabular}[c]{@{}c@{}}matched \\ tuple number\end{tabular}} & \multirow{2}{*}{\begin{tabular}[c]{@{}c@{}}top 1\\ number\end{tabular}} & \multirow{2}{*}{\begin{tabular}[c]{@{}c@{}}top 20\\ number\end{tabular}} & \multirow{2}{*}{\begin{tabular}[c]{@{}c@{}}top 50\\ number\end{tabular}} & \multirow{2}{*}{\begin{tabular}[c]{@{}c@{}}top 100\\ number\end{tabular}} \\ \cline{1-3}

\begin{tabular}[c]{@{}c@{}}embedding \\ category\end{tabular} & \begin{tabular}[c]{@{}c@{}}window\\ size\end{tabular} & \begin{tabular}[c]{@{}c@{}}embedding \\ size\end{tabular} & &  &  &  \\ \hline
\multirow{6}{*}{Glove} & 5 & 100 & 248 & 7 & 25 & 29 & 35 \\
 & 10 & 100 & 248 & 5 & 26 & 31 & 34 \\
 & 15 & 100 & 248 & 7 & 25 & 29 & 30 \\
 & 5 & 200 & 248 & 7 & 28 & 33 & 39 \\
 & 10 & 200 & 248 & 5 & 28 & 34 & 42 \\
 & 15 & 200 & 248 & 6 & 27 & 34 & 40 \\ \hline
\multirow{6}{*}{skip-gram} & 5 & 100 & 248 & 11 & 32 & 41 & 54 \\
 & 10 & 100 & 248 & 10 & 23 & 36 & 43 \\
 & 15 & 100 & 248 & 8 & 18 & 33 & 40 \\
 & 5 & 200 & 248 & 10 & 41 & 48 & 55 \\
 & 10 & 200 & 248 & 8 & 31 & 42 & 53 \\
 & 15 & 200 & 248 & \textbf{7} & 29 & 41 & 50 \\ \hline
\multirow{6}{*}{cbow} & 5 & 100 & 248 & 12 & 33 & 42 & 49 \\
 & 10 & 100 & 248 & 12 & 34 & 38 & 48 \\
 & 15 & 100 & 248 & 11 & 33 & 39 & 48 \\
 & 5 & 200 & 248 & 13 & 36 & 44 & 55 \\
 & 10 & 200 & 248 & 15 & 38 & 43 & 53 \\
 & 15 & 200 & 248 & 12 & 36 & 45 & 52 \\ \hline
\multirow{6}{*}{fastText} & 5 & 100 & 248 & 9 & 27 & 39 & 43 \\
 & 10 & 100 & 248 & 8 & 21 & 35 & 39 \\
 & 15 & 100 & 248 & 5 & 20 & 27 & 33 \\
 & 5 & 200 & 248 & 8 & 34 & 41 & 49 \\
 & 10 & 200 & 248 & 6 & 24 & 35 & 49 \\
 & 15 & 200 & 248 & 7 & 30 & 33 & 48 \\ \hline
\end{tabular}
}
\caption{Results of intrinsic metric by measuring the rank of similarity between word pairs in the range of top 1, 20, 50 and 100, using \textbf{simple wikipedia} as formal vocabulary. The minimum count of word for each word embedding setting is 200.}
\label{tb:res-simp}
\end{table*}

\begin{table*}[!ht]
\resizebox{\textwidth}{!}{\begin{tabular}{ccccccccc}
\hline
\multicolumn{3}{c|}{embedding} & \multirow{2}{*}{\begin{tabular}[c]{@{}c@{}}matched \\ tuple number\end{tabular}} & \multirow{2}{*}{\begin{tabular}[c]{@{}c@{}}top 1\\ number\end{tabular}} & \multirow{2}{*}{\begin{tabular}[c]{@{}c@{}}top 20\\ number\end{tabular}} & \multirow{2}{*}{\begin{tabular}[c]{@{}c@{}}top 50\\ number\end{tabular}} & \multirow{2}{*}{\begin{tabular}[c]{@{}c@{}}top 100\\ number\end{tabular}} \\ \cline{1-3}

\multicolumn{1}{c|}{\begin{tabular}[c]{@{}c@{}}embedding \\ category\end{tabular}} & \begin{tabular}[c]{@{}c@{}}window\\ size\end{tabular}  & \multicolumn{1}{c|}{\begin{tabular}[c]{@{}c@{}}embedding \\ size\end{tabular}} &  &  &  &  &  \\ \hline
\multicolumn{1}{c|}{\multirow{6}{*}{Glove}} & 5 & \multicolumn{1}{c|}{100} & 248 & 7 & 25 & 29 & 35 \\
\multicolumn{1}{c|}{} & 10  & \multicolumn{1}{c|}{100} & 248 & 5 & 26 & 31 & 34 \\
\multicolumn{1}{c|}{} & 15  & \multicolumn{1}{c|}{100} & 248 & 7 & 25 & 29 & 30 \\
\multicolumn{1}{c|}{} & 5 & \multicolumn{1}{c|}{200} & 248 & 7 & 28 & 33 & 39 \\
\multicolumn{1}{c|}{} & 10 & \multicolumn{1}{c|}{200} & 248 & 5 & 28 & 34 & 42 \\
\multicolumn{1}{c|}{} & 15 & \multicolumn{1}{c|}{200} & 248 & 6 & 27 & 34 & 40 \\ \hline
\multicolumn{1}{c|}{\multirow{6}{*}{skip-gram}} & 5 & \multicolumn{1}{c|}{100} & 248 & 11 & 32 & 41 & 55 \\
\multicolumn{1}{c|}{} & 10  & \multicolumn{1}{c|}{100} & 248 & 10 & 23 & 36 & 44 \\
\multicolumn{1}{c|}{} & 15  & \multicolumn{1}{c|}{100} & 248 & 8 & 18 & 33 & 40 \\
\multicolumn{1}{c|}{} & 5  & \multicolumn{1}{c|}{200} & 248 & 10 & 41 & 48 & 56 \\
\multicolumn{1}{c|}{} & 10 & \multicolumn{1}{c|}{200} & 248 & 8 & 31 & 42 & 53 \\
\multicolumn{1}{c|}{} & 15& \multicolumn{1}{c|}{200} & 248 & \textbf{7} & 29 & 41 & 50 \\ \hline
\multicolumn{1}{c|}{\multirow{6}{*}{cbow}} & 5 & \multicolumn{1}{c|}{100} & 248 & 13 & 27 & 34 & 56 \\
\multicolumn{1}{c|}{} & 10 & \multicolumn{1}{c|}{100} & 248 & 12 & 35 & 40 & 50 \\
\multicolumn{1}{c|}{} & 15 & \multicolumn{1}{c|}{100} & 248 & 11 & 33 & 40 & 50 \\
\multicolumn{1}{c|}{} & 5 & \multicolumn{1}{c|}{200} & 248 & 13 & 37 & 45 & 56 \\
\multicolumn{1}{c|}{} & 10  & \multicolumn{1}{c|}{200} & 248 & 15 & 39 & 44 & 54 \\
\multicolumn{1}{c|}{} & 15  & \multicolumn{1}{c|}{200} & 248 & 12 & 36 & 46 & 53 \\ \hline
\multicolumn{1}{c|}{\multirow{6}{*}{fastText}} & 5 & \multicolumn{1}{c|}{100} & 248 & 9 & 27 & 39 & 43 \\
\multicolumn{1}{c|}{} & 10 & \multicolumn{1}{c|}{100} & 248 & 8 & 21 & 35 & 39 \\
\multicolumn{1}{c|}{} & 15 & \multicolumn{1}{c|}{100} & 248 & 5 & 20 & 27 & 33 \\
\multicolumn{1}{c|}{} & 5& \multicolumn{1}{c|}{200} & 248 & 8 & 34 & 41 & 50 \\
\multicolumn{1}{c|}{} & 10 & \multicolumn{1}{c|}{200} & 248 & 6 & 25 & 31 & 51 \\
\multicolumn{1}{c|}{} & 15 & \multicolumn{1}{c|}{200} & 248 & 7 & 30 & 34 & 49 \\ \hline
\end{tabular}
}
\caption{Results of intrinsic metric by measuring the rank of similarity between word pairs in the range of top 1, 20, 50 and 100, using \textbf{English wikipedia} as formal vocabulary. The minimum count of word for each word embedding setting is 200.}
\label{tb:res-en}
\end{table*}

\begin{table*}[!h]
\resizebox{\textwidth}{!}{\begin{tabular}{c|cc|ccccc}
\hline
\multicolumn{3}{c|}{embedding} & \multirow{2}{*}{Epoch} & \multirow{2}{*}{\textbf{val\_acc}} & \multirow{2}{*}{\textbf{val\_loss}} & \multirow{2}{*}{acc} & \multirow{2}{*}{loss} \\ \cline{1-3}
\begin{tabular}[c]{@{}c@{}}embedding \\ category\end{tabular} & \begin{tabular}[c]{@{}c@{}}window\\ size\end{tabular} & \begin{tabular}[c]{@{}c@{}}embedding \\ size\end{tabular} &  &  &  &  &  \\ \hline
\multirow{6}{*}{Glove} & 5 & 100 & 8 & 0.9374 & 0.1953 & 0.9808 & 0.0709 \\
 & 10 & 100 & 14 & 0.9377 & 0.2330 & 0.9979 & 0.0181 \\
 & 15 & 100 & 12 & 0.9410 & 0.2292 & 0.9987 & 0.0114 \\
 & 5 & 200  & 11 & 0.9424 & 0.2158 & 0.9986 & 0.0121 \\
 & 10 & 200  & 15 & 0.9428 & 0.2501 & 0.9993 & 0.0056 \\
 & 15 & 200 & \textbf{13} & \textbf{0.9451} & 0.2276 & 0.9992 & 0.0077 \\ \hline
\multirow{6}{*}{skip-gram} & 5 & 100 & 14 & 0.9463 & 0.1881 & 0.9973 & 0.0233 \\
 & 10 & 100 & 11 & 0.9448 & 0.1846 & 0.9949 & 0.0317 \\
 & 15 & 100  & 9 & 0.9447 & 0.1836 & 0.9915 & 0.0437 \\
 & 5 & 200  & 12 & 0.9484 & 0.1952 & 0.9993 & 0.0064 \\
 & 10 & 200 & \textbf{14} & \textbf{0.9485} & 0.2036 & 0.9993 & 0.0077 \\
 & 15 & 200 & 8 & 0.9479 & 0.1754 & 0.9982 & 0.0174 \\ \hline
\multirow{6}{*}{cbow} & 5 & 100 & 6 & 0.9329 & 0.2314 & 0.9806 & 0.0639 \\
 & 10 & 100 & 8 & 0.9316 & 0.2697 & 0.9905 & 0.0346 \\
 & 15 & 100 & 9 & 0.9343 & 0.2749 & 0.9927 & 0.0266 \\
 & 5 & 200 & 10 & 0.9372 & 0.2671 & 0.9986 & 0.0089 \\
 & 10 & 200 & 9 & 0.9369 & 0.2733 & 0.9973 & 0.0129 \\
 & 15 & 200 & \textbf{7} & \textbf{0.9379} & 0.2510 & 0.9939 & 0.0234 \\ \hline
\multirow{6}{*}{fastText} & 5 & 100 & 11 & 0.9463 & 0.1764 & 0.9899 & 0.0475 \\
 & 10 & 100 & 14 & 0.9472 & 0.1965 & 0.9977 & 0.0195 \\
 & 15 & 100 & 14 & 0.9450 & 0.1844 & 0.9944 & 0.0337 \\
 & 5 & 200 & \textbf{12} & \textbf{0.9493} & 0.1867 & 0.9991 & 0.0112 \\
 & 10 & 200 & 6 & 0.9483 & 0.1639 & 0.9882 & 0.0527 \\
 & 15 & 200 & 11 & 0.9486 & 0.1849 & 0.9982 & 0.0174 \\ \hline
\end{tabular}
}
\caption{The maximum performance of our hashtag prediction model for each embedding, where column \textit{epoch} denotes at which epoch the model gains the maximum performance, the maximum performance on validation accuracy is marked as bold.}
\label{tb:pred_res}
\end{table*}

% en wiki
\begin{figure*}[!h]
% \vskip -5mm
\begin{center}
\centerline{\includegraphics[width=\columnwidth]{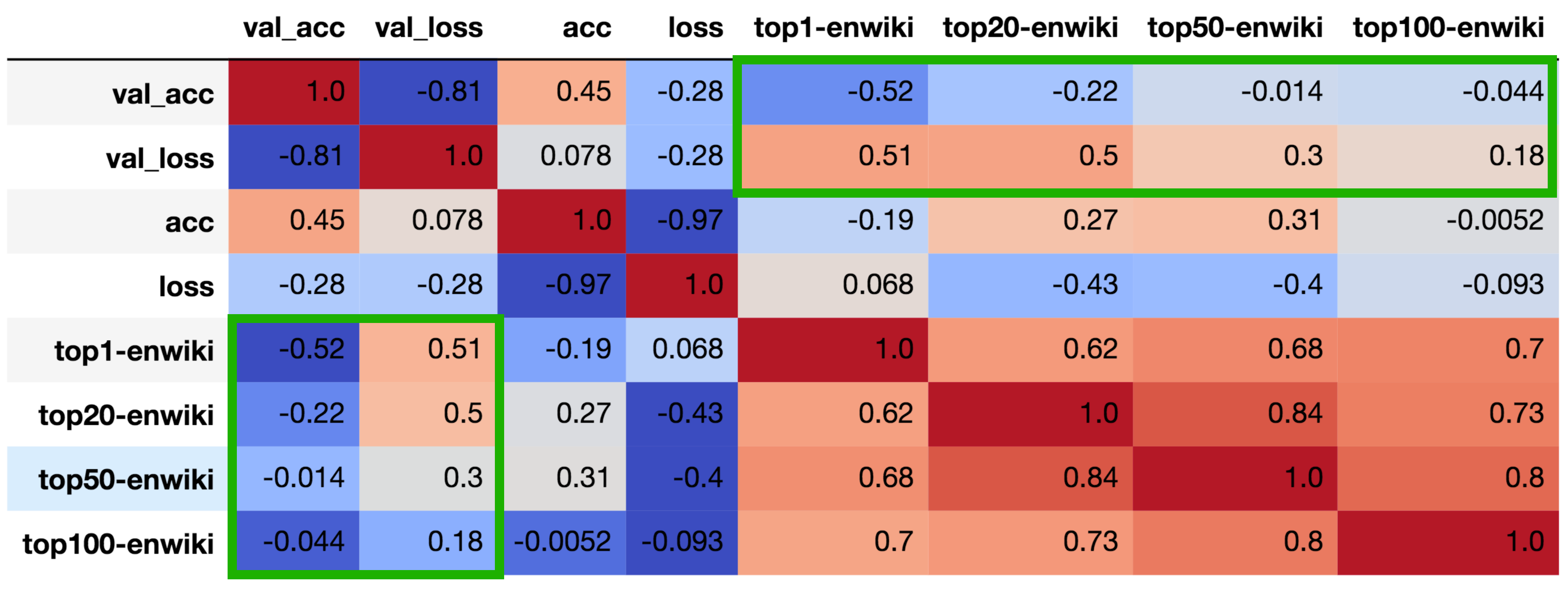}}
\caption{  \textbf{Pearson correlation} between the extrinsic model performance and the intrinsic metric using \textbf{English wikipedia} as the formal vocabulary.
}
\label{fig:pearson-en}
\end{center}
% \vskip -5mm
\end{figure*} 

% scatter
\clearpage
\begin{figure*}[!h]
\vskip -5mm
\begin{center}
\centerline{\includegraphics[width=\columnwidth]{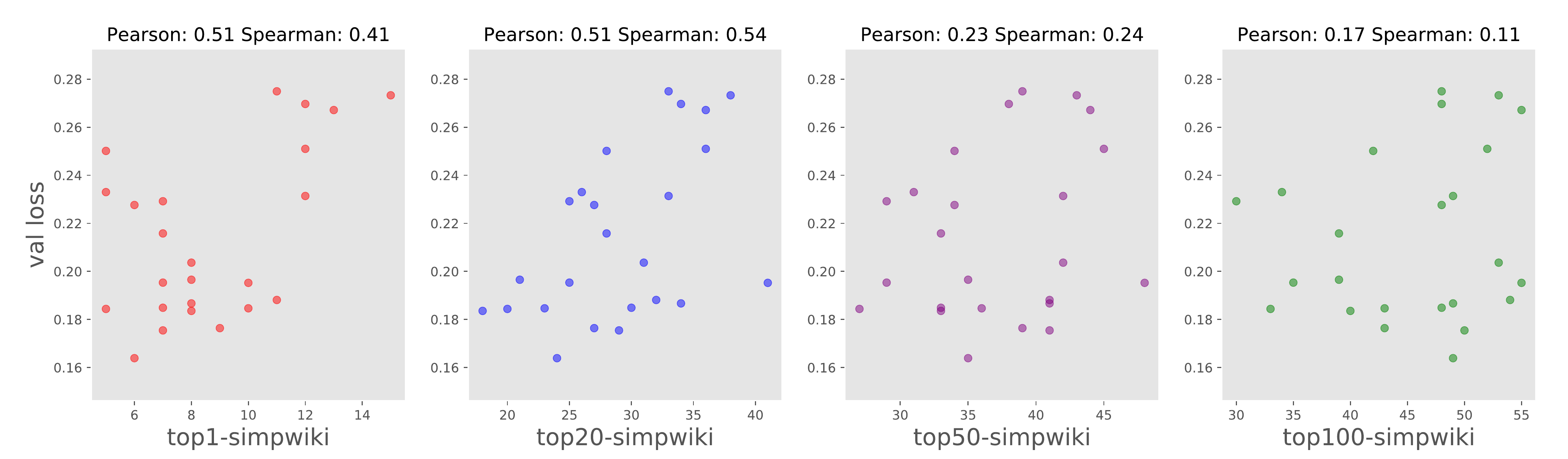}}
\caption{The scatter plot of the best \textbf{loss} on the validation set against the intrinsic metric performance using \textbf{simple wikipedia} as the formal vocabulary.
}
\label{fig:scatter-simp-loss}
\end{center}
\vskip -5mm
\end{figure*}

\begin{figure*}[!h]
\vskip -5mm
\begin{center}
\centerline{\includegraphics[width=\columnwidth]{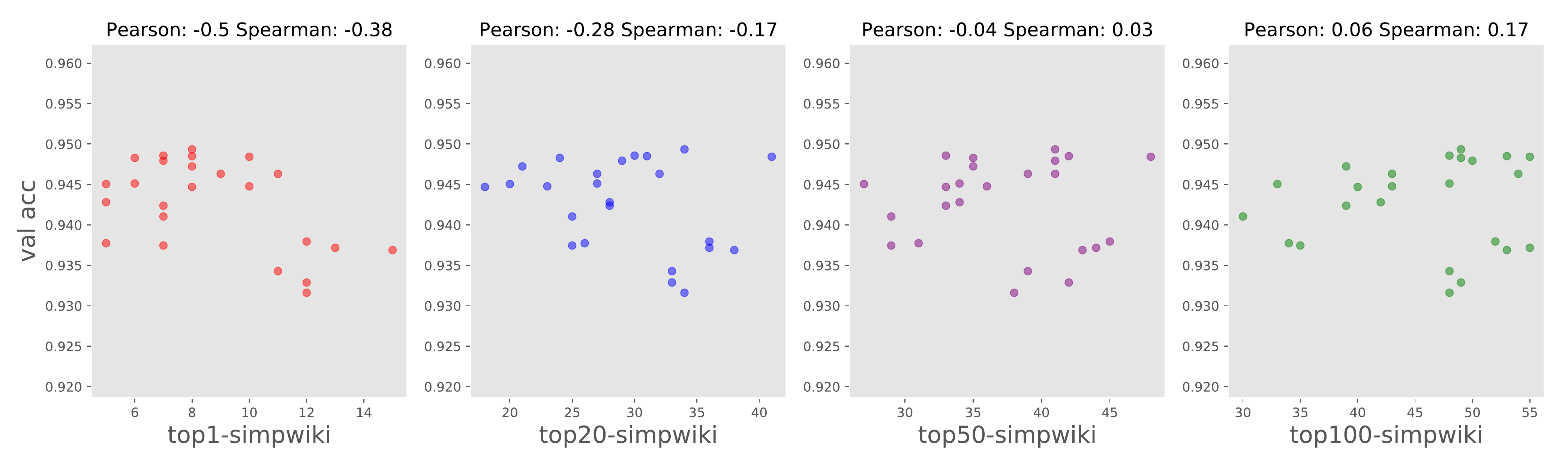}}
\caption{ The scatter plot of the best validation set \textbf{accuracy} against the intrinsic metric performance using \textbf{simple wikipedia} as the formal vocabulary.
}
\label{fig:scatter-simp-acc}
\end{center}
\vskip -5mm
\end{figure*}

\begin{figure*}[!h]
\vskip -5mm
\begin{center}
\centerline{\includegraphics[width=\columnwidth]{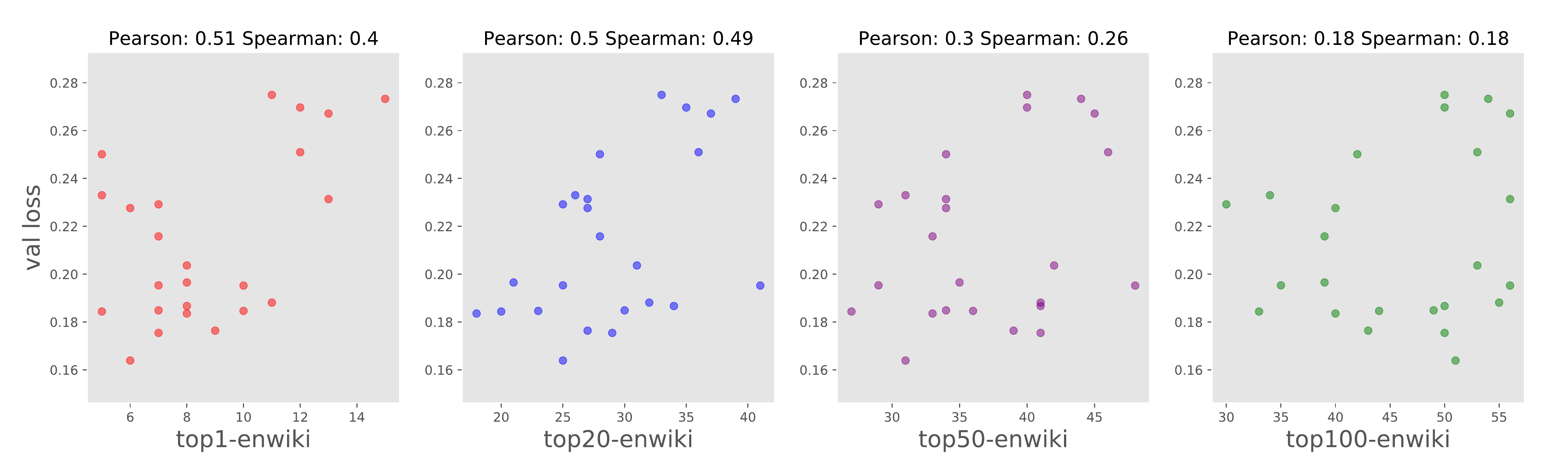}}
\caption{The scatter plot of the best \textbf{loss} on validation set against the intrinsic metric performance using \textbf{English wikipedia} as the formal vocabulary.
}
\label{fig:scatter-en-loss}
\end{center}
\vskip -5mm
\end{figure*}

\begin{figure*}[!ht]
\vskip -5mm
\begin{center}
\centerline{\includegraphics[width=\columnwidth]{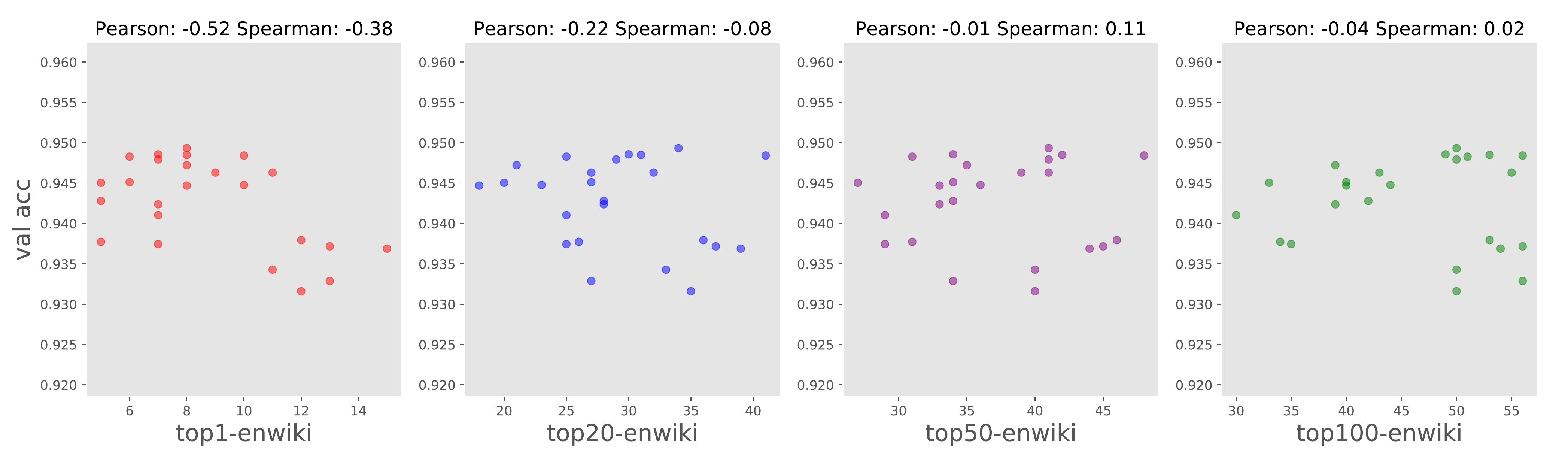}}
\caption{ The scatter plot of the best validation set \textbf{accuracy} against the intrinsic metric performance using \textbf{English wikipedia} as the formal vocabulary.
}
\label{fig:scatter-en-acc}
\end{center}
\vskip -5mm
\end{figure*}

\end{document}